\documentclass{article} 
\usepackage{iclr2019_conference,times}

\usepackage{hyperref}
\usepackage{url}

\usepackage[T1]{fontenc}
\usepackage[latin1]{inputenc}
\usepackage{subeqnarray}
\usepackage{amsfonts}
\usepackage{amsmath}
\usepackage{amssymb}
\usepackage{graphicx}
\usepackage{color}
\usepackage{stfloats}
\usepackage{balance}
\usepackage{epstopdf}
\usepackage{minibox}
\usepackage{xcolor,colortbl}
\usepackage{algpseudocode}
\usepackage{multirow}
\usepackage{hhline}

\usepackage[numbers]{natbib}

\title{Skeleton-based Gait Index Estimation with LSTMs}

\author{Trong-Nguyen Nguyen\\
DIRO, University of Montreal\\
Montreal, QC, Canada\\
\texttt{nguyetn@iro.umontreal.ca}\\
\And
Huu-Hung Huynh\\
University of Science and Technology\\
Danang, Vietnam\\
\texttt{hhhung@dut.udn.vn}\\
\And
Jean Meunier\\
DIRO, University of Montreal\\
Montreal, QC, Canada\\
\texttt{meunier@iro.umontreal.ca}\\
}

\iclrfinalcopy 
\begin{document}

\maketitle

\begin{abstract}
In this paper, we propose a method that estimates a gait index for a sequence of skeletons. Our system is a stack of an encoder and a decoder that are formed by Long Short-Term Memories (LSTMs). In the encoding stage, the characteristics of an input are automatically determined and are compressed into a latent space. The decoding stage then attempts to reconstruct the input according to such intermediate representation. The reconstruction error is thus considered as a weak gait index. By combining such weak indices over a long-time movement, our system can provide a good estimation for the gait index. Our experiments on a large dataset (nearly one hundred thousand skeletons) showed that the index given by the proposed method outperformed some recent works on gait analysis.
\end{abstract}

\section{Introduction}
Vision-based health-care systems are nowadays becoming popular because of the fast development of related research fields such as computer vision and machine learning. The task of gait analysis is one application example that has been focused on with such systems. Many researchers attempted to deal with this problem using various input data types such as subject silhouette~\cite{Bauckhage2005,Bauckhage2009}, skeleton~\cite{Paiement2014,Tao2016,Nguyen2016}, depth map~\cite{Rougier2011}, typical color image~\cite{Andriluka2009}, and a combination of types~\cite{Nguyen2018BHI}. These inputs usually require different pre-processing operations to have a reasonable representation. In our work, we select the skeleton as the input of our system since it is already represented by a collection of 3D body joints and thus does not need any complicated pre-processing as the others. This stage is typically followed by a feature extraction, in which the gait features are defined under spatial and/or temporal aspects and are thus usually interpretable. Differently from such studies, we perform this task automatically by structuring our index estimator as auto-encoders, where the features of interest are salient  inside the network without any supervision. An auto-encoder consists of an encoder, that converts the input into an appropriate representation in a latent space, and a decoder that attempts to reconstruct the input based on the salient features. The difference between an input and its output is thus a reasonable choice as a (weak) gait index. Concretely, by fitting an auto-encoder using patterns of specific gait types, the loss between an input and its reconstruction should be useful to indicate a (normalized or non-normalized) probability that the input belongs to such known gait types.

An auto-encoder is a special type of neural networks, where the input and output are pair-wise defined. Instead of using a feed-forward network, we design our model as a Recurrent Neural Network (RNN) with LSTM cells. There are several reasons for this selection. First, a RNN can embed the temporal factor into its weights. This is an important property because many studies on gait analysis (e.g.~\cite{Bauckhage2009,Paiement2014,Tao2016,Nguyen2016}) demonstrated that temporal gait assessment provided higher accuracies than assessing individually each frame. Second, a RNN does not require a large capacity for storing parameters compared with non-cyclic neural networks that can provide the same accuracy. Third, a RNN can deal with variable-length inputs while a feed-forward network only works on inputs of a fixed size. Although we performed our experiments (see Section~\ref{sec:experiments}) on inputs of the same size, this property is also important in practical applications where the skeletons may be acquired under different walking speeds or camera frame rates.

The remaining of this paper is organized as follows: Section~\ref{sec:method} describes the structure details as well as the workflow of our system; the employed dataset and our experimental results with the proposed method together with some related works are presented in Section~\ref{sec:experiments}; and Section~\ref{sec:conclusion} gives the conclusion.

\section{Method} \label{sec:method}

\subsection{Skeletal input}
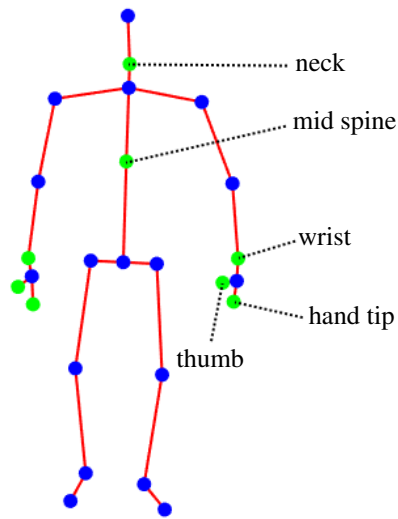
\begin{figure}[t]
\centering
\begin{picture}(150,197)
	\put(0,0){\includegraphics[scale=0.82]{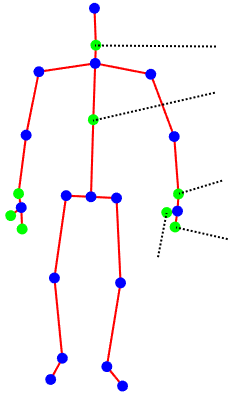}}
	\put(110,170){neck}
	\put(109,148){mid spine}
	\put(111,104){wrist}
	\put(65,58){thumb}
	\put(115,74){hand tip}
\end{picture}
\caption{An illustration of our joint selection. The blue circles indicate the 17 selected joints and the green circles correspond to the 8 discarded joints.}
\label{fig:skeleton}
\end{figure}

As mentioned in the previous section, the input of our system is a sequence of skeletons. In our work, we employ a Microsoft Kinect 2 to determine 3D skeletons based on captured depth maps using an existing functionality~\cite{Shotton2011realtime}. Each skeleton is represented by a collection of 25 joint positions in a 3D space.

We do not feed a sequence of the entire 25 joints into the model. Instead, a simple joint selection and a normalization are performed as a skeletal data enhancement. Concretely, there are 8 body joints that are discarded from each input skeleton due to their lack of efficiency in describing the posture for gait analysis. The joint selection is illustrated in Fig.~\ref{fig:skeleton}. The neck joint is not considered because this is nearly an interpolation of the head and the shoulder spine. The collection of wrist, hand tip and thumb belonging to each body side can be considered as the elements of a cluster that is represented by the corresponding hand joint. Besides, these 3 joints do not have a significant contribution on the task of gait analysis. Finally, the mid spine is also discarded because of its lack of movement freedom compared with the others. The input then becomes a sequence of 17 joint coordinates in 3D space.

The next step in our processing is to split the input into 3 sequences of 17 scalar values corresponding to the 3 axes. Beside simplifying the input, this task is also convenient to determine the dependence of gait analysis accuracy on each space axis. Finally, the data range along each axis is normalized by scaling it into [0, 1]. This step synchronizes the input range, our model can thus deal with skeletons that perform at different distances from the camera. The input sequence of skeletons is now three sequences of vectors, where each one contains 17 values stretching from 0 to 1. Notice that we use 3 auto-encoders for the three sequences.

\subsection{Model structure}
\begin{figure}[t]
\centering
\begin{picture}(250,230)
	\put(0,70){\includegraphics[width=0.634\textwidth]{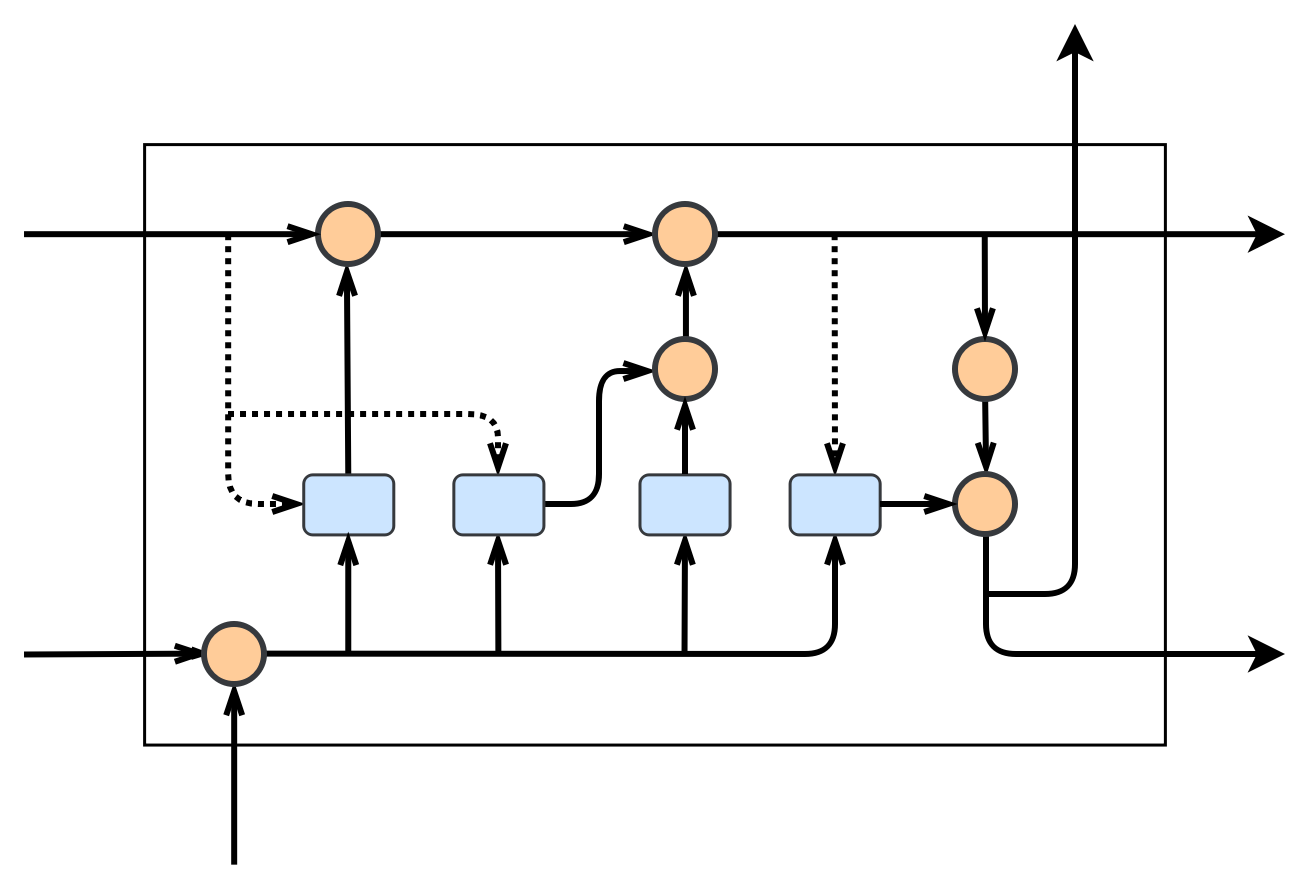}}
	\put(70,0){\includegraphics[scale=0.3]{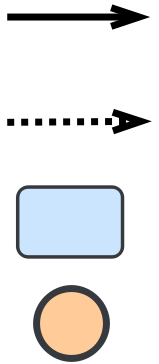}}
	\put(110,61){data transfer}
	\put(110,42){peephole connection}
	\put(110,23){network layer}
	\put(110,4){element-wise operation}
	\put(62.5,194){$\times$} \put(127.8,194){$+$}
	\put(127.5,168){$\times$} \put(186,167){$\delta$}
	\put(63.6,142){$\sigma$} \put(93,142){$\sigma$}
	\put(128.3,141.2){$\delta$} \put(157,142){$\sigma$}
	\put(185.4,142){$\times$}
	\put(42.1,113.3){$\bullet$}
	\put(5,203){$c_{t-1}$}\put(3,121){$h_{t-1}$}\put(48,85){$x_t$}
	\put(190,220){$h_t$}\put(226,121){$h_t$}\put(229,203){$c_t$}
	\put(68,175){$f_t$}\put(107,170){$i_t$}\put(134,154){$\hat{c}_t$}
	\put(172,150){$o_t$}
\end{picture}
\caption{The LSTM architecture with peephole connections. The $\bullet$ operation indicates a concatenation.}
\label{fig:LSTM}
\end{figure}
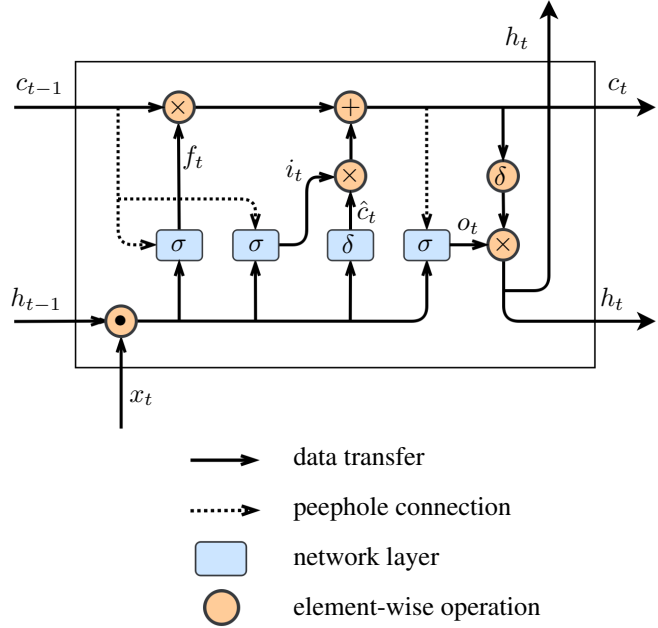

\begin{figure}[t]
\centering
\begin{picture}(260,95)
	\put(0,3){\includegraphics[width=0.634\textwidth]{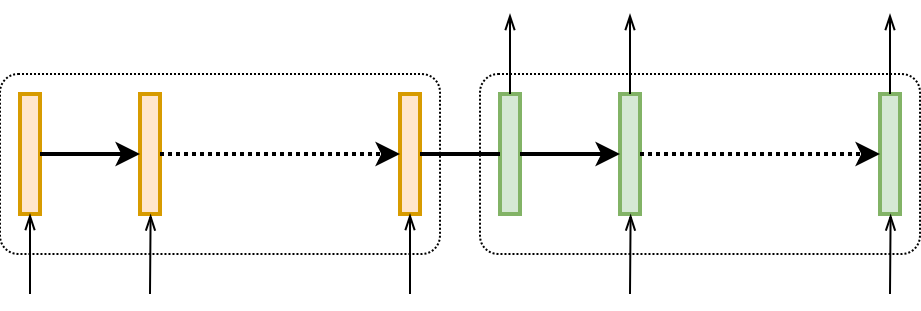}}
	\put(60,30){encoder}\put(191,30){decoder}
	\put(3,0){$x_1$}\put(36,0){$x_2$}\put(108,0){$x_T$}
	\put(167,0){$y_T$}\put(240,0){$y_2$}
	\put(134,90){$y_T$}\put(163,90){$y_{T-1}$}
	\put(240,90){$y_1$}
\end{picture}
\caption{Our auto-encoder that uses two LSTMs for the encoding and decoding stages. An output $y_t$ is the reconstruction of its input $x_t$.}
\label{fig:autoencoder}
\end{figure}
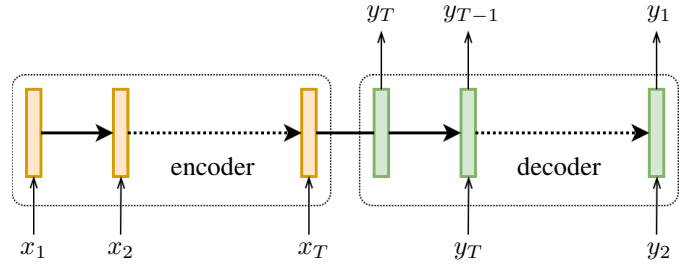

Our system is formed by auto-encoders where each one can be separated into an encoder and a decoder that share the same latent space. We design each of the two parts as a one-block RNN with an individual set of weights. In this work, the LSTM with peephole connections~\cite{Gers2003} is employed to represent that block (Fig.~\ref{fig:autoencoder}). This LSTM structure has been applied successfully in many applications such as speech recognition~\cite{Graves2013}, large scale acoustic modeling~\cite{Hasim2014}, language modeling~\cite{Sundermeyer2012}, and video representation~\cite{Srivastava2015}.

Figure~\ref{fig:LSTM} presents the overview of a LSTM structure. The strength of a LSTM comes from the cell state $c_t$ that contains the information and the gates that control how such information should be transferred. Concretely, a LSTM has 3 possible inputs including the current input $x_t$, the cell state $c_{t-1}$ and the output $h_{t-1}$ of the previous loop. The forget gate applies a sigmoid activation $\sigma$ on such inputs to decide (by $f_t$) which information would be removed from the cell state. The input gate then selects the partial information needing to be updated by a sigmoid output $i_t$. The update is performed according to this decision together with new values $\hat{c}_t$ that are provided by a tanh layer $\delta$. The new cell state $c_t$ finally goes through a tanh activation to calculate the output $h_t$ of the current loop with the support of the output gate that provides $o_t$. The variables $c_t$ and $h_t$ are transferred to the next loop. These operations can be formulated and ordered as follows.
\begin{equation}
	i_t = \sigma\big(W_{ix}x_t + W_{ih}h_{t-1} + W_{ic}c_{t-1} + b_i\big),~~~~~
	\label{eq:i}
\end{equation}
\begin{equation}
	f_t = \sigma\big(W_{fx}x_t + W_{fh}h_{t-1} + W_{fc}c_{t-1} + b_f\big),~~~
	\label{eq:f}
\end{equation}
\begin{equation}
	\hat{c}_t=\delta\big(W_{\hat{c}x}x_t + W_{\hat{c}h}h_{t-1} + b_{\hat{c}}\big),~~~~~~~~~~~~~~~~~~~~~~~~
	\label{eq:c_hat}
\end{equation}
\begin{equation}
	c_t = f_t \ast c_{t-1} + i_t \ast \hat{c}_t,~~~~~~~~~~~~~~~~~~~~~~~~~~~~~~~~~~~~~~~~
	\label{eq:c}
\end{equation}
\begin{equation}
	o_t = \sigma\big(W_{ox}x_t + W_{oh}h_{t-1} + W_{oc}c_t + b_o\big),~~~~~~~~~
	\label{eq:o}
\end{equation}
\begin{equation}
	h_t = o_t \ast \delta\big(c_t\big)~~~~~~~~~~~~~~~~~~~~~~~~~~~~~~~~~~~~~~~~~~~~~~~~~~~~~~~~
	\label{eq:h}
\end{equation}
where $\ast$ indicates an element-wise multiplication, $W_{uv}$ is the weight matrix connecting the gate $u$ and the data source $v$, in which the weights of peephole connections $W_{ic}$, $W_{fc}$ and $W_{oc}$ are diagonal matrices, $\sigma$ and $\delta$ are respectively the sigmoid and tanh functions.

The LSTM used in our encoder and the one in the decoder have the same architecture but different parameters (e.g. weight matrices). An overview of our unrolled auto-encoder is shown in Fig.~\ref{fig:autoencoder}. The input of this model is a sequence of $x_t$ where $1\le t\le T$. The encoder attempts to compress this input and then transfer the result to the decoder. In the decoding stage, the model outputs the reconstructions of $x_t$ one by one in the reverse order. In this work, we empirically used 256 hidden units for each LSTM.

\subsection{Gait index estimation}
A simple gait index can be defined as a loss between an input sequence $\{x_1,x_2,...,x_T\}$ and its reconstruction $\{y_1,y_2,...,y_T\}$ that is outputted from the auto-encoder. Recall that each sequence of skeleton gives 3 sequences of 17-element vectors. Therefore we use 3 auto-encoders for such 3 inputs. We expect that the optimization of a model focusing on a specific space axis would be easier than such task on a mixture of 3 axes. Therefore, we create and independently train the 3 models. A simple gait index can thus be estimated as a combination of the three reconstruction errors.

In this work, we use the Mean Square Error (MSE) as the loss function. The MSE corresponding to each auto-encoder is also used as a weak gait index. Since we need to combine these 3 indices to have a better measure for the input sequence of skeletons, a simple sum (or average) seems to be appropriate. However, the weak indices provided by the 3 models may contribute to a reasonable gait index with various amounts. Therefore, we decided to use a weighted sum to estimate the desired index. We determine the weight of each model according to its MSE that is calculated over the training set. Since the MSE indicates how bad the output sequence has been reconstructed, a model with a high error value should have a small weight and vice versa. The weight $w_k$ of a model $k$ $(k\in\{X,Y,Z\})$ with the corresponding MSE $e_k$ is estimated as
\begin{equation}
	w_k=e_k^{-1}\sum_ke_k
	\label{eq:weight}
\end{equation}
This equation satisfies the inverse relationship between a training error and its weight. The sum of training errors is put into eq.~(\ref{eq:weight}) to keep the resulting weight inside a reasonable value range even if $e_k$ is very small or very large.

\section{Experiments} \label{sec:experiments}
In order to assess our proposed gait index, we performed experiments and evaluated the results according to a popular application of such index. Concretely, we focus on the task of detecting abnormal walking gaits. An efficient gait index estimator must assign indices to normal and abnormal gaits so that these two ensembles can be well separated using an appropriate threshold. Some recent works also employed this task for their evaluation such as~\cite{Paiement2014,Tao2016,Nguyen2016}.

\subsection{Dataset}
In our experiments, we used a dataset that was acquired by 9 subjects that walked on a treadmill at a speed of 1.28 kph. Each subject performed a normal walking gait and 8 abnormal ones. The gaits with abnormality were simulated by padding a sole (of 3 possible thicknesses: 5, 10, 15 centimeters) under one foot or attaching a weight (4 kilograms) to one ankle. A Kinect 2 was placed in front of the subject to capture frontal-view silhouettes and skeletons. The data corresponding to each gait of a subject were represented by 1200 consecutive frames. The silhouette was employed in our experiments to evaluate another related work in order to provide a comparison.

We separated the dataset (i.e. silhouettes as well as skeletons) into two smaller sets. The first one was employed as a training set that includes the gaits of 5 subjects, and the test set consisting of the remaining gaits of 4 volunteers.

\subsection{Index estimation from coarse to fine}
\begin{figure}[t]
\centering
\begin{picture}(250,65)
	\put(0,8){\includegraphics[width=0.634\textwidth]{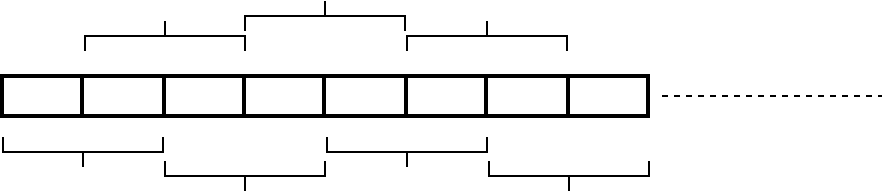}}
	\put(19,7){$s_1$}\put(66,0){$s_3$}\put(112,7){$s_5$}\put(158,0){$s_7$}
	\put(43,62){$s_2$}\put(88,68){$s_4$}\put(135,62){$s_6$}
\end{picture}
\caption{The overlapping split of each normal gait sequence in the training set. The term $s$ indicates a sequence of 12 skeletons and is an input of our system in the training stage.}
\label{fig:split}
\end{figure}
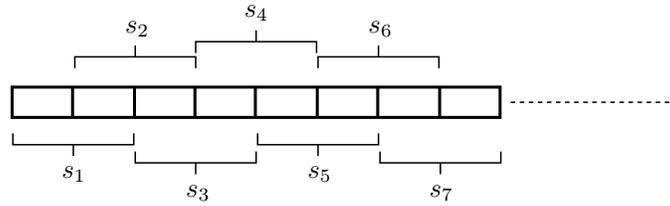

In order to perform the task of abnormal gait detection, we used only normal gaits to train the three weak index estimators. An input of abnormal gait is expected to provide a bad reconstruction compared with normal gait. The gait index, i.e. the MSE between an input and its reconstruction, can thus be easily employed to distinguish between normal and abnormal gaits. Each normal gait sequence of 1200 skeletons in the training set was split into overlapping sub-sequences of length $T = 12$ where two consecutive ones sharing 6 skeletons. This step is illustrated in Fig.~\ref{fig:split}. The use of overlapping was to increase the number as well as the diversity of training samples. Each auto-encoder was thus trained with 995 inputs for 5 subjects in the training set. Notice that in the test phase, we separated gait sequences into consecutive non-overlapping segments. The indices estimated by each of the three models as well as the weighted sum are per-segment indices since they are measured on a short sequence of skeletons (of length 12 in our experiments). In order to compute the index for each entire sequence of 1200 skeletons in our test set, we simply calculated the mean value of the indices corresponding to consecutive short segments.

\begin{figure}[t]
\centering
\includegraphics[width=0.7\textwidth]{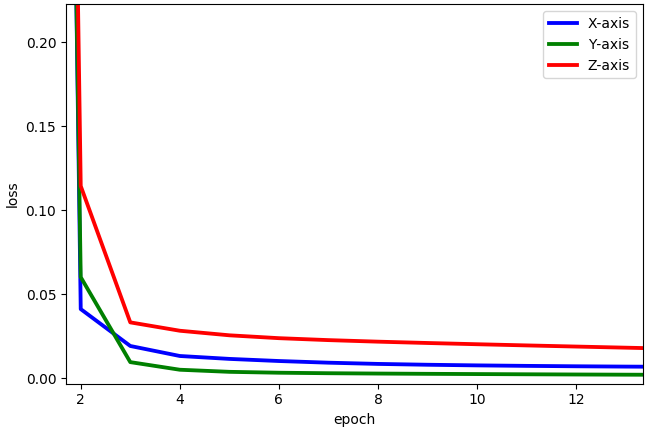}
\caption{Training loss of the three auto-encoders.}
\label{fig:trainingloss}
\end{figure}

The three models were trained with 100 epochs and the model loss during the training stage is shown in Fig.~\ref{fig:trainingloss}.
It was obvious that these models quickly converged after just a few training epochs. It also showed that the convergence speed of the Z-axis auto-encoder was slower than the two others, thus its resulting index might be less efficient. In an effort to enhance the proposed system, we added a dropout layer~\cite{Hinton2014} to the connection between the model input and the encoder LSTM as suggested in~\cite{Zaremba2014}. The use of dropout is to reduce the risk of overfitting but requires more training iterations.

An important factor in our experiments is the evaluation quantity. Instead of using just a typical classification accuracy, we employed the Area Under Curve (AUC) of the Receiver Operating Characteristic (ROC) curve to indicate the efficiency of our proposed gait index. The AUC measures how well a collection of indices belonging to two groups can be separated according to an appropriate threshold. This quantity has also been used for evaluation in many studies dealing with binary classification problems (e.g.~\cite{Tao2016,Nguyen2016}). Another advantage of the ROC curve is that it can measure an Equal Error Rate (EER) that is comparable with the typical classification accuracy. In our experiments, we also measured the EER together with related measures such as precision, specificity, sensitivity, accuracy, and F1-score. The experimental results are presented in Table~\ref{table:results}.
\begin{table*}[t]
\centering
\caption{Experimental results of our approach from coarse to fine (top-down).}
\label{table:results}
\vspace{5pt}
\scriptsize
\begin{tabular}{|l|l||ccccccc|}
\hline
\multicolumn{2}{|c||}{\textbf{Index estimation}}  & \textbf{AUC} & \textbf{EER} & \textbf{Sensitivity} & \textbf{Specificity} & \textbf{Precision} & \textbf{Accuracy} & \textbf{F1-score} \\ \hline \hline
\multirow{6}{*}{per-segment} & X-axis model & 0.883 & 0.185 & 0.815 & 0.815 & 0.972 & 0.815 & 0.887 \\ 
    & Y-axis model & 0.863 & 0.223 & 0.777 & 0.777 & 0.965 & 0.777 & 0.861 \\ 
    & Z-axis model & 0.695 & 0.356 & 0.644 & 0.645 & 0.936 & 0.644 & 0.763 \\ \cline{2-9} 
    & non-weighted sum & 0.805 & 0.268 & 0.732 & 0.733 & 0.956 & 0.732 & 0.829 \\ 
    & weighted sum & 0.902 & 0.172 & 0.828 & 0.830 & 0.975 & 0.828 & 0.896 \\
		& weighted sum + dropout & 0.910 & 0.167 & 0.833 & 0.833 & 0.975 & 0.833 & 0.899 \\ \hline
\multirow{3}{*}{per-sequence} & non-weighted sum & 0.844 & 0.278 & 0.719 & 0.750 & 0.958 & 0.722 & 0.821 \\ 
    & weighted sum & 0.953 & 0.083 & 0.906 & 1.000 & 1.000 & 0.917 & 0.951 \\
    & weighted sum + dropout& \textbf{0.969} & \textbf{0.056} & \textbf{0.938} & \textbf{1.000} & \textbf{1.000} & \textbf{0.944} & \textbf{0.968} \\ \hline
\end{tabular}
\end{table*}
\begin{figure*}[t]
\centering
\scalebox{0.765}{
\begin{picture}(520,185)
	\put(0,20){\includegraphics[scale=0.51]{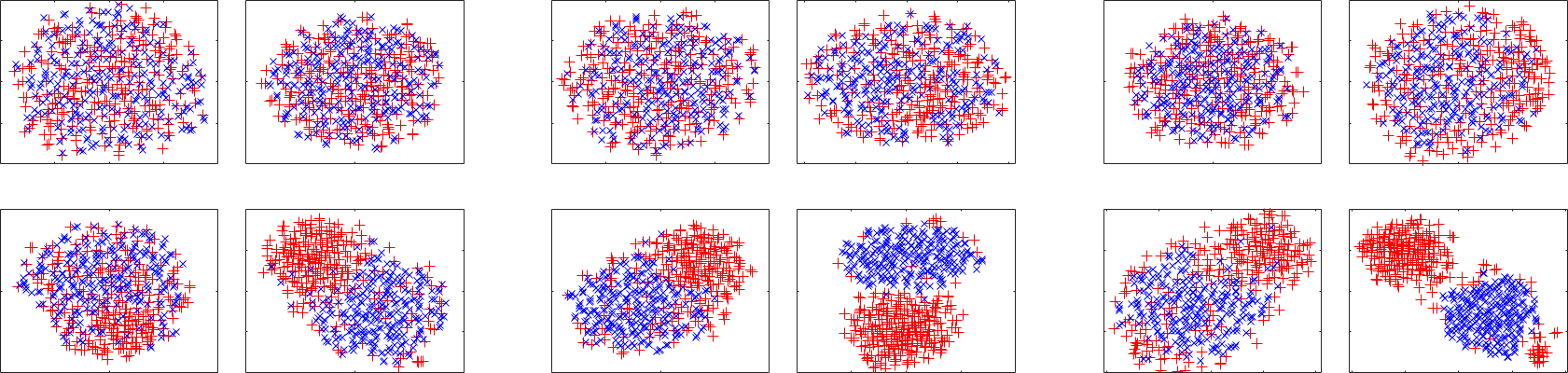}}
	\put(26,0){(a) X-axis auto-encoder}
	\put(209,0){(b) Y-axis auto-encoder}
	\put(392,0){(c) Z-axis auto-encoder}
	\put(5,148){\small $W_{ix}$ (input gate)}
	\put(108,148){\small $W_{\hat{c}x}$}
	\put(3,79){\small $W_{fx}$ (forget gate)}
	\put(83,79){\small $W_{ox}$ (output gate)}
	\put(186,148){\small $W_{ix}$ (input gate)}
	\put(289,148){\small $W_{\hat{c}x}$}
	\put(184,79){\small $W_{fx}$ (forget gate)}
	\put(264,79){\small $W_{ox}$ (output gate)}
	\put(367,148){\small $W_{ix}$ (input gate)}
	\put(470,148){\small $W_{\hat{c}x}$}
	\put(365,79){\small $W_{fx}$ (forget gate)}
	\put(445,79){\small $W_{ox}$ (output gate)}
	\put(182,165){\includegraphics[scale=0.89]{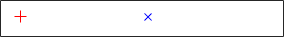}}
	\put(199,173.5){\small with dropout}
	\put(267,173.5){\small without dropout}
\end{picture}}
\caption{Visualizing distributions of LSTM weights that are related to input connections using t-SNE~\cite{Maaten2008}.}
\label{fig:weights}
\end{figure*}
The weak index provided by the Z-axis model was less efficient than the two others as we guessed according to Fig.~\ref{fig:trainingloss}. The gait index estimated by the weighted combination was also significantly better than using a simple sum. By adding a dropout layer (where the probability of retention was 0.5 in our experiments), the system was slightly enhanced. Finally, estimating the gait index over a long sequence was better than considering short segments. Therefore, the index of a segment could be considered as a partial index that may contain noise, and a smoothed version of such indices is an enhancement of gait index representation.

In order to know how the system was enhanced when the dropout was added, we looked at the LSTM weights related to input connections in the encoder. Figure~\ref{fig:weights} visualizes the weights of 256 hidden units: $W_{ix}$, $W_{fx}$, $W_{\hat{c}x}$ and $W_{ox}$ in eq.~(\ref{eq:i}),~(\ref{eq:f}),~(\ref{eq:c_hat}) and~(\ref{eq:o}), respectively. It was apparent that the distributions of the weight at the input gate ($W_{ix}$) and the one that supported estimating new candidate values of the cell state ($W_{\hat{c}x}$) were almost unchanged when adding the dropout. The distributions of the two other weights, that connected the input with the forget and output gates, tended to expand and move their centers when adding a dropout layer, especially for the latter gate. Therefore, the dropout seemed to improve the mapping from input space to latent space by modifying the forget and output gates.

\subsection{Comparison with related methods}
In order to compare our approach with some related studies that employed hand-crafted features, we reimplemented the works~\cite{Bauckhage2009} and~\cite{Nguyen2016}. The inputs of the system~\cite{Bauckhage2009} are subject's silhouettes captured from a frontal view. The gait classification can be performed on each frame with/without considering recent frames using a binary Support Vector Machine (SVM) as well as over a sequence using a trigger. The study~\cite{Nguyen2016}, similarly to ours, requires a sequence of skeletons as the input. The gait assessment can be performed on each walking gait cycle using a Hidden Markov Model (HMM) and over the entire sequence using a non-linear computation. We reimplemented a HMM for~\cite{Nguyen2016} and a one-class SVM for~\cite{Bauckhage2009} since our work focuses on the unsupervised learning. This consideration is appropriate for practical applications because collecting samples of abnormality with enough generalization is quite difficult. A comparison of these experimental results and ours is presented in Table~\ref{table:comparison}.
\begin{table*}[t]
\centering
\caption{Experimental results of related works.}
\label{table:comparison}
\vspace{5pt}
\begin{tabular}{l||lcc}
\hline
\multirow{2}{*}{\textbf{Model}} & \multirow{2}{*}{\textbf{Data type}} & \multicolumn{2}{c}{\textbf{Classification error}}\\ \cline{3-4} 
 & & per-segment & per-sequence \\ \hline \hline
HMM~\cite{Nguyen2016} & skeleton & 0.335 & 0.250 \\ \hline
One-class SVM~\cite{Bauckhage2009} & silhouette & 0.227 & 0.139 \\ \hline
Ours (weighted sum) & skeleton & 0.172 & 0.083 \\
Ours (weighted sum + dropout) & skeleton & \textbf{0.167} & \textbf{0.056} \\ \hline
\end{tabular}
\end{table*}
The task of distinguishing between normal and abnormal gaits provided the best results (both per-segment and per-sequence) when employing the proposed index estimation. Our results were still better than the studies~\cite{Bauckhage2009} and~\cite{Nguyen2016} even when detaching the dropout.

\section{Conclusion} \label{sec:conclusion}
This paper proposes an approach for gait index estimation that requires a sequence of skeletons. Unlike related studies, the feature extraction is implicitly performed using auto-encoders. By employing a LSTM for the encoder and another one for the decoder, our system has the ability to work with temporal inputs. A weak gait index is estimated by joint coordinates belonging to each 3D axis. A weighted combination of the three weak indices significantly improves the efficiency of our index estimation. By adding a dropout layer right after the input, our system is slightly enhanced and outperformed related studies that work with skeletons as well as silhouettes.

\subsubsection*{Acknowledgment}
The authors would like to thank the NSERC (Natural Sciences and Engineering Research Council of Canada) for having supported this work (Discovery Grant RGPIN-2015-05671). This work was also supported by The Ministry of Education and Training, Vietnam, Grant KYTH-59.

\bibliography{references}
\bibliographystyle{iclr2019_conference}

\end{document}